\documentclass[conference]{IEEEtran}
\IEEEoverridecommandlockouts
\usepackage{cite}
\usepackage{amsmath,amssymb,amsfonts}
\usepackage{algorithmic}
\usepackage{graphicx}
\usepackage{textcomp}
\usepackage{xcolor}
\usepackage{booktabs}
\def\BibTeX{{\rm B\kern-.05em{\sc i\kern-.025em b}\kern-.08em
    T\kern-.1667em\lower.7ex\hbox{E}\kern-.125emX}}
\begin{document}

\title{Multi-task CNN Behavioral Embedding Model For Transaction Fraud Detection \\
}

\author{
\IEEEauthorblockN{1\textsuperscript{st} Bo Qu}
\IEEEauthorblockA{\textit{eBay, Decision Sciences} \\
Shanghai, China \\
boqu@ebay.com}
\and
\IEEEauthorblockN{ }
\IEEEauthorblockA{ }
\and
\IEEEauthorblockN{2\textsuperscript{nd} Zhurong Wang}
\IEEEauthorblockA{\textit{eBay, Decision Sciences} \\
Austin, USA \\
zhurowang@ebay.com}
\and
\IEEEauthorblockN{ }
\IEEEauthorblockA{ }
\and
\IEEEauthorblockN{3\textsuperscript{rd} Minghao Gu}
\IEEEauthorblockA{\textit{eBay, Decision Sciences} \\
Austin, USA \\
mingu@ebay.com}
\and
\IEEEauthorblockN{ }
\IEEEauthorblockA{ }
\and
\IEEEauthorblockN{4\textsuperscript{th} Daisuke Yagi}
\IEEEauthorblockA{\textit{eBay, Decision Sciences} \\
Houston, USA \\
dyagi@ebay.com}
\and
\IEEEauthorblockN{ }
\IEEEauthorblockA{ }
\and
\IEEEauthorblockN{5\textsuperscript{th} Yang Zhao}
\IEEEauthorblockA{\textit{eBay, Risk Engineering} \\
Shanghai, China \\
yzhao5@ebay.com}
\and
\IEEEauthorblockN{ }
\IEEEauthorblockA{ }
\and
\IEEEauthorblockN{6\textsuperscript{th} Yinan Shan}
\IEEEauthorblockA{\textit{eBay, Risk Engineering} \\
Shanghai, China \\
yshan@ebay.com}
\and
\IEEEauthorblockN{ }
\IEEEauthorblockA{ }
\and
\IEEEauthorblockN{7\textsuperscript{th} Frank Zahradnik}
\IEEEauthorblockA{\textit{eBay, Decision Sciences} \\
Austin, USA \\
francis.zahradnikiii@ebay.com}
}

\maketitle

\begin{abstract}
The burgeoning e-Commerce sector requires advanced solutions for the detection of transaction fraud. With an increasing risk of financial information theft and account takeovers, deep learning methods have become integral to the embedding of behavior sequence data in fraud detection. However, these methods often struggle to balance modeling capabilities and efficiency and incorporate domain knowledge. To address these issues, we introduce the multitask CNN behavioral Embedding Model for Transaction Fraud Detection. Our contributions include 1) introducing a single-layer CNN design featuring multirange kernels which outperform LSTM and Transformer models in terms of scalability and domain-focused inductive bias, and 2) the integration of positional encoding with CNN to introduce sequence-order signals enhancing overall performance, and 3) implementing multitask learning with randomly assigned label weights, thus removing the need for manual tuning. Testing on real-world data reveals our model's enhanced performance of downstream transaction models and comparable competitiveness with the Transformer Time Series (TST) model.
\end{abstract}


\section{Introduction}

The exponential growth of e-Commerce has led to a surge in online transactions, with fraud detection becoming increasingly crucial to mitigate substantial financial losses. The sophisticated landscape of e-Commerce calls for robust and efficient fraud detection methods capable of handling complex real-time transactional data. In particular, multivariate time series (MTS) have gained traction, illuminating patterns of customer behavior over time, aiding in identifying fraud.

LSTM \cite{karim2019multivariate} and Transformers \cite{zerveas2021transformer}\cite{chen2019behavior} are widely used deep learning methods for the classification of MTS. Nevertheless, they have limitations including the LSTM's increased computational requirements, Transformers' high memory consumption, and their tendency to overlook critical information by analyzing the entire sequence before generating predictions. Moreover, most conventional models focused on single-task learning, resulting in missed opportunities to capture shared features among various fraudulent activities faced by e-Commerce business. Additionally, real-world transaction dataset complexity requires improved strategies for integrating different feature types within a unified framework.

In this paper, we present the Multi-task CNN (MTCNN) Behavioral Embedding Model for Transaction Fraud Detection. This model outperforms LSTM and Transformers in computational scalability and captures risk behavior within short time periods. It adopts a multitask learning approach, mitigating overfitting issues, improving generalization capabilities, and eliminating manual label weight adjustment. In addition, we have incorporated positional encoding with CNN to introduce sequence-order signals, which has led to performance boost. Furthermore, we integrate a scaling embedding method for continuous variables in MTS data and combines it with standard categorical variable embeddings for a holistic representation, which is a universal embedding method to help our real word data integrate easier to other MTS models. In real-world industry data, MTCNN exhibits improved performance compared to other models that use transformers, such as the TST model \cite{zerveas2021transformer}.

The remainder of this paper discusses related work in multivariate time series modeling and multitask learning, details of the proposed methodology, experiments on real-world data, results, and potential future research directions.

\section{Related Work}

The scope of this paper encompasses two broad areas: multivariable time series modeling and multitask learning.

\subsection{MTS modeling}

MTS have been effectively managed by deep learning, with LSTM and Transformer models emerging as notable tools to capture complex temporal dependencies \cite{ismail2019deep}. Notable LSTM models include Multivariate LSTM-Fully Convolutional Network (MLSTM-FCN) \cite{karim2019multivariate}, which combined with attention mechanisms, has excelled in action recognition. TapNet \cite{zhang2020tapnet} integrates traditional time series approaches with deep learning for efficient low-dimensional feature learning, albeit struggling with limited labeled data. Deep learning coupled with the Markov transition field (MTF) has shown promise in fraud detection, with RNN-based structures enhancing predictive capacity \cite{zhang2018sequential}, but computational intensity remains a concern. Transformer models like TST \cite{zerveas2021transformer} and Behavior Sequence Transformer \cite{chen2019behavior} have also demonstrated effectiveness in MTS but suffer from computational and scalability challenges due to quadratic parameter growth.

CNN have proven to be highly effective in modeling MTS, outperforming both LSTM networks and Transformers. One reason for this success is their superior scalability - they require less memory than other models and can easily be parallelized. In addition, CNNs use a specialized \lq kernel\rq\space mechanism that allows them to quickly identify anomalous or fraudulent behaviors over short periods of time. Key advancements include Fully Convolutional Network (FCN) \cite{wang2017time}, Multiscale Convolutional Neural Network (MCNN) \cite{cui2016multi}, Multichannel Deep Convolutional Neural Networks (MC-DCNN) \cite{zheng2016exploiting}, Multivariate Convolutional Neural Network (MVCNN) \cite{liu2018time}, and MTEX-CNN \cite{assaf2019mtex}, each introducing unique strategies to improve accuracy, efficiency, and explainability in MTS. Despite their success, CNNs often struggle with feature learning and handling the sequential nature of data. In this study, we propose a CNN architecture combined with embedding techniques and positional encoding that addresses these challenges while being computationally scalable and lightweight.

\subsection{Enhancing Generalizability with Multi-task Learning}

Multitask learning (MTL) \cite{vandenhende2021multi} has gained traction to enhance the generalizability of machine learning models, particularly in transaction fraud detection. The two main MTL architectures are hard parameter sharing, exemplified by UberNet \cite{kokkinos2016ubernet}, and soft parameter sharing, represented by Cross-Stitch Networks \cite{Misra_2016_CVPR}. While hard parameter sharing is favored for simplicity and reduced overfitting, soft parameter sharing can be more robust but may increase the architectural complexity. Task balance methods such as Uncertainty Weights (UW) \cite{kendall2018multi}, Gradient Normalization (GradNorm) \cite{chen2018gradnorm}, and Random Loss Weighting (RLW) \cite{lin2022reasonable} offer strategic approaches to MTL optimization. Our design choice combines the simplicity and reduced overfitting of hard parameter sharing and the generalization ability of RLW in the Multitask Convolutional Neural Network (MT-CNN) structure, providing an effective approach to transaction fraud detection.

\begin{figure*}[ht]
\centering
\includegraphics[width=\textwidth]{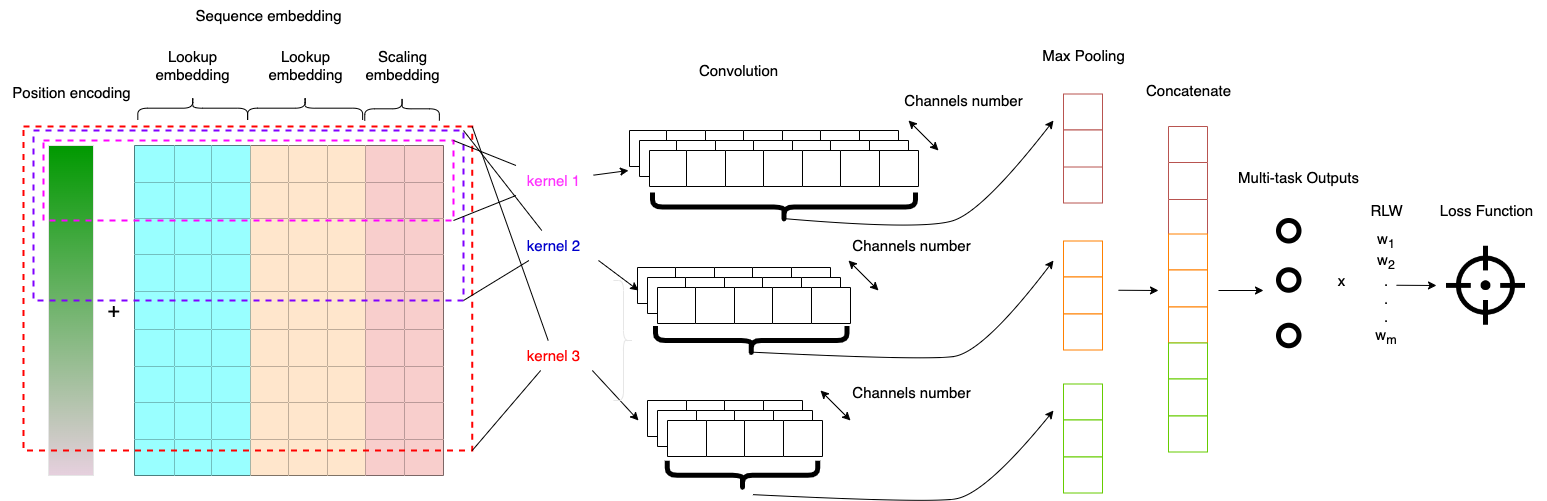}  
\caption{MTCNN model architecture with three channels as example.}
\label{fig:mtcnn}
\end{figure*}

\section{Approach}

 As shown in Figure \ref{fig:mtcnn}, in this study, we propose a robust and scalable framework for transaction fraud detection employing a one-layer Convolutional Neural Network (CNN) architecture with multi-range kernels, integrating multitask learning for correlated fraud-related tasks, and using a random label weight scheme to balance the contribution of different tasks. Additionally, a novel approach to variable embeddings for MTS data is introduced, focusing on handling continuous variables effectively and involving events' order information. The methodology section is structured as follows: first, we present our MTS data notation and embedding methodology, followed by a detailed explanation of our architecture, multitask learning approach, random label weight integration, and model optimization and evaluation methods.

\subsection{MTS Data Notation and Embedding}

\begin{figure*}[ht]
\centering
\includegraphics[width=\textwidth]{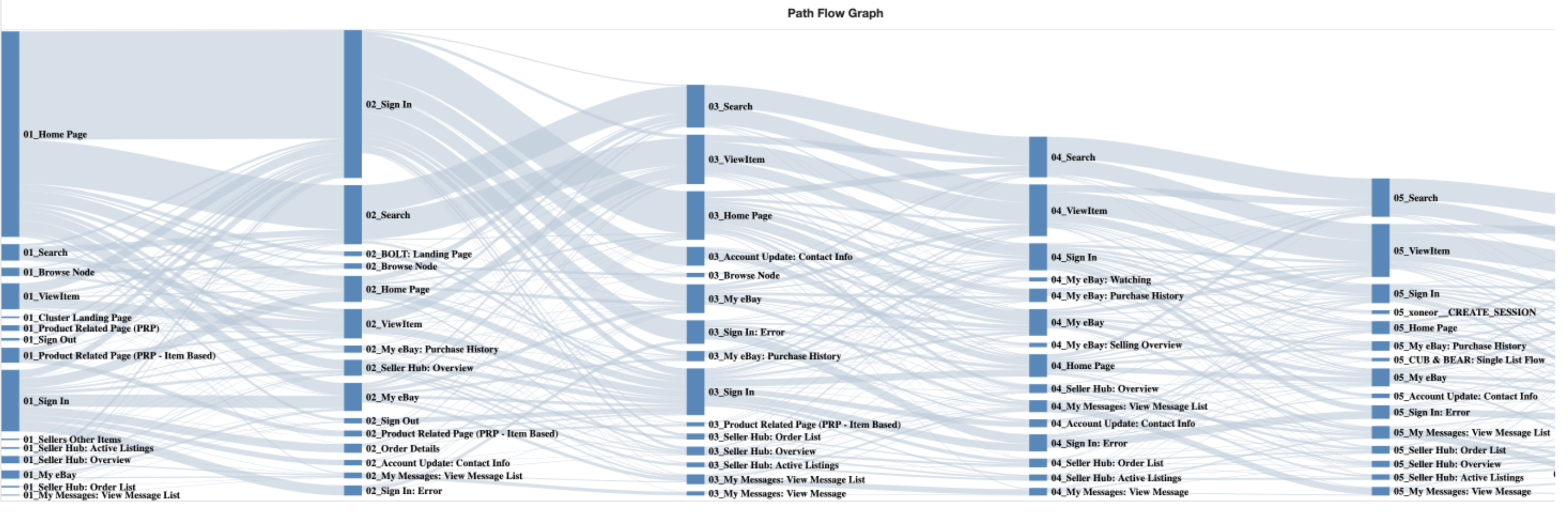}  
\caption{Illustration of user behavior flow of page browsing.}
\label{fig:behavior_flow}
\end{figure*}

Our framework is applied to MTS data, specifically user page view sequences from an e-commerce website. Each sequence represents a user's browsing behavior and includes three variables per time step: page ID, item category, and page view time. As depicted in Figure \ref{fig:behavior_flow}, the page ID signifies the specific page the user visited, the item category represents the type of items displayed on the page, and the page view time measures the user's dwell time on that page.

Formally, an MTS is denoted as $S=\{s_1, s_2,...,s_L\}$, where the length of time series $L$ can vary in real-world applications. Each sequence step $s_i = \{p_i, c_i, t_i\}$, with $p_i$ denoting the page ID, $c_i$ indicating the category of the item, and $t_i$ representing the page view time. In our study, we used padding and truncation techniques to preprocess all MTS samples to a uniform length $N$. This length, $N$, is adjustable based on the specific requirements of sequence modeling.

The page ID and item category variables are categorical, and they are preprocessed through label encoding, while rare categories are filtered and missing values are assigned a specific constant. The page view time is a continuous variable and is logarithmically normalized as $t_i' = \log(t_i/1000)$ to account for its wide distribution. 

Our model employs different embedding methods for categorical and continuous variables. For categorical variables, we adopt a traditional approach and use an embedding lookup table with a dimension size specified for each variable. The embedding weights are initialized using the Xavier uniform method. Continuous variables, on the other hand, are processed differently. We use a single embedding lookup for the continuous variable, and the actual continuous value is multiplied by the embedding to obtain the final embedded representation. 

Formally, let us denote the embedding lookup function for a variable $v$ as $E_v(\cdot)$, the discrete variables as $d=\{p_i,c_i\}$, and the continuous variable as $t_i'$. The embeddings of the discrete variables are $E_d(p_i)$ and $E_d(c_i)$, while the embedding of the continuous variable $t_i'$ is $t_i' \times E_{t'}(0)$, where $\times$ denotes the multiplication of elements in the element. 

The combined embedding of $s_i$ is thus a concatenation of the above three embeddings, given as: 

\begin{equation}
S_i' = \left[E_d(p_i); E_d(c_i); t_i' \times E_{t'}(0)\right]
\end{equation}

These embedded representations capture more detailed and richer information about the page view sequences, thus equipping our model with more granular insights for detecting transaction fraud.

\subsection{Positional Encoding}

Inspired by \cite{vaswani2017attention,devlin2019bert}, as with sequences in NLP tasks, the order in which events occur is an important factor in understanding the dynamics of user behavior in e-commerce systems. Thus, we introduce a positional encoding (PE) mechanism into the model, enabling the injection of sequential information into the data. This is particularly crucial in our task, as user behaviors are generally sequential, and capturing such dependencies can significantly improve the model's predictive capabilities.

In the spirit of flexibility and experimentation, our architecture supports two types of positional encoding: Fixed Positional Encoding (FPE) and Learnable Positional Encoding (LPE). Both types aim to inject information about the absolute position of the tokens in the sequence. For fixed PE, we use sinusoidal functions of different frequencies, in line with the original transformer architecture \cite{vaswani2017attention}. For the LPE, on the other hand, we utilized a learnable random parameter matrix that has the same dimension as FPE, which allows the model to learn and adjust the positional embeddings during the training process. In our experiments, both methods yielded similar results, suggesting that the additional complexity of learnable embeddings may not be necessary for our task.

We add positional encoding to the embedded sequence input $s_i'$ by element-wise addition:

\begin{equation}
s_i'' = s_i' + PE(s_i'),
\end{equation}

where $PE(.)$ denotes the positional encoding operation.

\subsection{Convolutional Neural Network}

The core of our architecture is a one-layer CNN with multi-range kernels. Following the embedding of the MTS data, we draw inspiration from the field of Natural Language Processing (NLP) \cite{chen2015convolutional, zhang2015sensitivity}. The encoded sequence $s_i''$ is fed into a series of convolutional layers, each equipped with a different kernel size to capture patterns of various lengths. Mathematically, given a set of kernel sizes $K = {k_1, k_2, ..., k_n}$, for each kernel size $k_j \in K$, the convolution operation can be represented as:

\begin{equation}
C_{k_j} = Conv_{k_j}(s_i''),
\end{equation}

where $Conv_{k_j}(.)$ denotes the convolution operation with kernel size $k_j$. Each of these operations results in a feature map $C_{k_j}$.

Following the convolution, we apply batch normalization to standardize the feature maps and then perform a max-pooling operation over time. The max-pooling operation aims to capture the most significant features across the sequence. The resulting tensor is then passed through a Rectified Linear Unit (ReLU) activation function for nonlinearity.

Finally, the output of all convolutional layers is concatenated along the feature dimension, forming a unified feature representation. Mathematically, this operation can be denoted as:

\begin{equation}
F = [C_{k_1}; C_{k_2}; ...; C_{k_n}],
\end{equation}

where $[;]$ denotes the concatenation operation.

The concatenated feature tensor is then passed through a fully-connected layer, followed by a dropout operation for regularization. Afterward, another fully connected layer transforms the features into the final output tensor $o$ with the desired dimensionality, that is, the number of classes for the classification task. Thus, our model can be expressed as follows.

\begin{equation}
o = FC2(Dropout(ReLU(FC1(F)))),
\end{equation}

where $FC1(.)$ and $FC2(.)$ denote the first and second fully connected layers, respectively.

\subsection{Multi-task Learning}

We integrate MTL with our architecture to improve the generalization capability and predictive performance of the model. Specifically, we utilize a traditional hard parameter sharing framework and implement RLW for multitask weighting, ensuring a balanced contribution from various tasks throughout the learning process.

Given the CNN output of one layer $o$, the model predicts several output tasks with the same shared parameters, which can be represented as:

\begin{equation}
o_{i, task} = W_{task}(o),
\end{equation}

where $o_{i, task}$ is the output tensor for the task $i$ using the shared parameters and $W_{task}(.)$ denotes the task-specific weight computation.

The use of hard parameter sharing is preferable because our tasks share a significant amount of features due to their correlation in fraud detection scenarios. For example, unauthorized chargebacks, stolen financial information, and account takeovers share common features, and hard parameter sharing can help to learn a more generalized embedding space across these tasks.

To optimize the model in an MTL setting, we employ the RLW optimization strategy. The RLW optimization generates a set of softmax-normalized weights of the tasks, W, by sampling from a normal distribution at each training iteration:

\begin{equation}
W = \frac{\exp(\xi)}{\sum_{i=1}^{n}\exp(\xi_i)},
\end{equation}

where $n$ is the number of tasks and $\xi$ is a random variable following a normal distribution. This dynamic adjustment of the weights enables the model to adapt to variances in the importance of different tasks.

Using the weights $W$, we compute the cross-entropy (CE) loss for each task as follows:

\begin{equation}
L_{CE}^{task} = -\sum_i y_i^{task} \log(\hat{y}_i^{task}),
\end{equation}

where $y_i^{task}$ represents the ground truth and $\hat{y}_i^{task}$ is the predicted probability for the task $i$. The overall loss function is then constructed using a weighted combination of the task-specific losses:

\begin{equation}
L = \sum_{task=1}^{T} W_{task} L_{CE}^{task},
\end{equation}

where $T$ denotes the total number of tasks.

 RLW optimization is driven by its unique capability to dynamically adjust the weights for each training iteration, effectively balancing the contribution of different tasks. This is particularly beneficial for our application, as certain tasks may be more critical than others depending on the specifics of a transaction, but it is hard to know the exact correlations and find an optimized task weight manually. By continuously generating different random weights at each training iteration, RLW optimization enables the model to adapt to these variances and consistently provide optimal predictions. According to the RLW article, this method is shown to have a higher probability of escaping local minima, resulting in better generalization ability and achieving comparable performance with state-of-the-art baselines on various benchmark datasets.

\subsection{Architecture Advantages}

Our architecture, which integrates a simplified CNN structure with positional encoding and MTL, offers several advantages. Firstly, the computational scalability of CNNs makes it suitable for handling large-scale datasets, as in many e-Commerce applications. Furthermore, unlike more complex architectures such as LSTMs and transformers, our simplified design reduces the risk of overfitting, particularly when the amount of available data is limited.

Secondly, the multi-range kernels allow our model to capture both short-term and long-term behavioral patterns, a feature that is highly desirable in many practical applications. In addition, the positional encoding infuses sequence information into the data, enabling the model to effectively capture the sequential nature of user behavior in e-Commerce systems.

Lastly, the incorporation of MTL improves the generalizability and predictive performance of the model by allowing the extraction of shared features among related tasks in fraud detection scenarios. The hard parameter sharing technique employed in our MTL approach facilitates the learning of a more generalized embedding space between tasks. Furthermore, we also applied an RLW optimization strategy that dynamically adjusts task weights to better adapt to variances in the importance of different tasks during the learning process.

In conclusion, our methodology presents a flexible and efficient approach to the task of detecting risk behavior, striking a balance between model complexity and computational efficiency while retaining the ability to capture the essential temporal, contextual, and multivariate information of user behavior.

\section{Experiment}
In our study, we performed the experiments in two stages. During the first stage, we collected buyer transaction data and trained two models: our proposed MTCNN model and our customized version of the TST model \cite{zerveas2021transformer}, which is a representative approach that takes advantage of the powerful transformer architecture for MTS. Both models were trained using the same multitask supervised labels and their outputs were compared on the basis of the Kolmogorov-Smirnov (KS) statistic and the Information Value (IV). In the second stage, we gathered more buyer transaction data and features and trained Gradient Boosted Machine (GBM) based Transaction Fraud detection models. The outputs derived from the MTCNN and TST models were incorporated as additional features, allowing us to gauge the relative performance improvements offered by the MTCNN model compared to the TST model by analyzing the performance of downstream models that utilize their outputs as additional features.

\subsection{Datasets and Evaluation Metrics}
To evaluate our proposed method, we sampled real-world transaction data for both stages of the experiments. Multiple types of fraudulent transactions were utilized as labels to assess the multitask performance of our model's output. These fraudulent transactions were classified into different risk categories, such as financial theft and account takeover, among others. Three primary buyer transaction fraud types were involved: Task 1, Task 2, and Task 3. Fraudulent transactions were labeled 1, while legitimate transactions were labeled 0. The dataset was chronologically divided into training and testing sets. Since legitimate transactions are much more than the fraudulent transaction, we applied random downsampling for legitimate transactions.

The features used in the first stage of the experiment were sequences of user behavior, as described in Section 3.1. In the second stage of the experiment, the features included those derived from the data warehouse feature tables, which were already employed in Buyer transaction risk models. These features included velocity features, minimum/maximum/average aggregation features, time difference features, and categorical features, among others. Moreover, the outputs of the first stage sequence embedding models were incorporated as additional features; these outputs consisted of scores and vectors from the final hidden layer of our sequence embedding models. Detailed information on the data set can be found in Table \ref{tab:DatasetInfo}.

For evaluation metrics, we used the KS statistic and IV to measure the predictive power of the output in the first stage. In the second stage, we employed dollar amount weighted Precision-Recall Area Under the Curve (\$PR-AUC) and dollar amount weighted recall (\$r) at specific precision (\$p) points to assess the performance of the downstream models that incorporated the outputs from our sequence embedding models.

\begin{table}
\caption{Experiment Datasets}
\centering
\resizebox{\columnwidth}{!}{
\begin{tabular}{|c|c|c|c|c|c|c|}
\hline
Experiments & Dataset  & Total   & Task 1 Fraud & Task 2 Fraud & Task 3 Fraud \\
\hline
Stage 1 & Train  & 7588890 &29811 &16097 &6561     \\ 
                           & Test  & 4412299 &12081 &6731 &3014     \\
\hline
Stage 2 & Train  &12400377 &99023 &59495 &13449     \\ 
                           & Test   &5350661 &31137 &15087 &4714     \\
\hline
\end{tabular}
}
\label{tab:DatasetInfo}
\end{table}

\subsection{Experimental Results and Analysis}

\paragraph{Stage 1}
For the first stage of the experiments, we trained two models using the dataset described above: our proposed MTCNN model and the TST model. The input behavior sequences consisted of three variables for each timestamp: page ID, item category, and page view time. After preprocessing as described in Section 2.4.1, we set the maximum sequence length to 100 and applied padding and truncation operations. Regarding the parameters of the MTCNN model, we manually explored several options, such as kernel sizes, channel numbers, and positional encoding methods. Ultimately, we opt for kernel sizes of $[8, 16, 32, 64]$, channel numbers of $[50, 50, 50, 50]$, fixed positional encoding, and a dropout rate of 0.5. For optimization, we used Adam Optimizer with a learning rate of 1e-4.

Concerning the parameters of the TST model, we followed the guidelines of the original paper to fully utilize the transformer architecture. We involved a self-learning pretraining strategy combined with multitask supervised fine-tuning, which proved to be more effective than multitask supervised training from scratch. We chose $2$ transformer layers with $4$ attention heads and adopted BatchNorm and the GELU activation function as suggested by the original paper. Additionally, during self-learning pretraining, we applied a random masking process and state transition probabilities for masked segments with lengths that adhere to a geometric distribution, as recommended in the paper. Unlike the original paper, during supervised fine-tuning, we experimented with different representation pooling layers, including max pooling, mean pooling, max-mean pooling, and CNN one-layer pooling. In practice, CNN pooling of one layer performed comparatively better. This layer used three kernels with sizes $[8,16,32]$ and channel numbers of $50$ for each kernel. During fine-tuning, we did not freeze the pre-trained parameters, mirroring the original paper's approach. The optimization method and parameters were comparable to the MTCNN model.

Following training, we evaluated the outputs of the MTCNN and TST models using KS and IV scores for each task label. Table \ref{tab:Stage1Readout} lists the models' output scores for KS and IV, highlighting MTCNN's competitive performance even when compared to a representative method like TST. This shows that the CNN architecture remains an effective choice for MTS models.
\begin{table}
 \caption{KS/IV Comparison between MTCNN and MTTST}
  \centering
  \begin{tabular}{lllll}
    \toprule
     &\multicolumn{2}{c}{MTCNN} & \multicolumn{2}{c}{MTTST}                     \\
    \midrule
    Scores & KS   &IV  & KS   & IV  \\
    \midrule
    
    Task 1 score &\textbf{53.8} &\textbf{2.014} & 52.02 &1.943   \\ 
    Task 2 score &\textbf{64.76} &\textbf{2.832} & 63.54 &2.775   \\ 
    Task 3 score &\textbf{63.33} &\textbf{2.531} & 62.55 &2.457   \\ 
    \bottomrule
  \end{tabular}
  \label{tab:Stage1Readout}
\end{table}

\paragraph{Stage 2}
With the trained models from Stage 1, we prepared additional training data for Stage 2, as outlined previously. We inferred the sequence models' outputs on the training data as additional features. The GBM method used for training was a modified version of MTGBM \cite{ying2022mtgbm} custom made by us, which we adapted modifications for our multitask transaction fraud detection scenario. Three experiments were conducted in this stage: training an MTGBM model with original tabular features and training two additional MTGBM models with additional features from MTCNN and TST, respectively. Feature selections, such as PSI $<$ 0.2, were applied to the added features of the sequence models. The MTGBM parameters were identical for each experiment, using default values such as learning rate 0.03, max\_bin 20, max\_depth 7, number of leaves 80, bagging fraction 0.9, and feature\_fraction 0.8.

We evaluated the outputs of the three models in a subset of testing data, naming the Universal Control Group (UCG), using the \$PR-ROC metric and \$recall at specific \$precision points. The UCG data constitute approximately 10\% of the overall test transactions shown in Table \ref{tab:DatasetInfo}. Almost no actual measures were implemented to obstruct dubious transactions, making it ideal for measurement purposes, as all genuine fraudulent transactions were preserved for assessment. Also note that the data used in Stage 2 were obtained six months after the Stage 1 model training, so the result will be less exaggerated than using the dataset close to Stage 1. In addition, because the data size for our Stage 2 experiment is considerably large, we opted to train the MTGBM models once for each of the three models using consistent, adequately effective parameters based on prior experience, as we expect the standard deviation to be fairly low. 

As depicted in Table \ref{tab:Stage2Readout}, consistent with the KS / IV results in stage 1, MTCNN showcased a competitive performance boost compared to the TST method. For the \$PR-ROC we observed superior results from TST, but in the actual \$recall at specific \$precision comparisons, which are more important for the real world model applications in transaction fraud rules, MTCNN still have good performance compared to TST. These findings indicate CNN's enhanced stability and robust generalization capabilities when working with real-world data, particularly considering that CNN is easier to be deployed in Near Real Time (NRT) fraud detection system than Transformer models. Furthermore, these findings also indicate that there could be additional benefits in combining models built by different architectures for different usage scenarios, since from the \$PR-ROC results, our customized TST model also has potential advantages in other Precision/Recall ranges.

\begin{table}
 \caption{ Performance of MTGBM with/without MTCNN and TST output features in Multiple Tasks, on UCG Test Data.}
  \centering
  \begin{tabular}{llll}
    \toprule
    \multicolumn{4}{c}{Task 1}                   \\
    \midrule
    Experiments & \$PR-AUC   &\$r@\$p $\approx$ 25\%  &\$p@\$r$\approx$80\%\\
    \midrule
    MTGBM      &0.5939           &87.9\%          &40.95\%           \\ 
    MTGBM+MTCNN &\textbf{0.6044} &\textbf{89.4\%} &\textbf{48.95\%}  \\ 
    MTGBM+MTTST   &0.6029          &88.72\%         &47.36\%           \\ 
    \bottomrule
    \toprule
    \multicolumn{4}{c}{Task 2} \\
    \midrule
    Experiments & \$PR-AUC   &\$r@\$p$\approx$25\%  &\$p@\$r$\approx$80\%  \\
    \midrule
    MTGBM             &0.5876           &88.37\%         &39.93\%   \\ 
    MTGBM+MTCNN       &0.5922           &\textbf{90.5\%} &\textbf{44.81\%}  \\ 
    MTGBM+MTTST         &\textbf{0.5934}  &90.15\%          &43.8\%  \\ 
    \bottomrule
    \toprule
    \multicolumn{4}{c}{Task 3} \\
    \midrule
    Experiments & \$PR-AUC   &\$r@\$p$\approx$25\%  &\$p@\$r$\approx$80\%  \\
    \midrule
    MTGBM      &0.3179          &67.06\%          &14.06\%                             \\ 
    MTGBM+MTCNN &0.3671       & \textbf{81.86\%} &\textbf{25.65\%}                                                 \\ 
    MTGBM+MTTST   &\textbf{0.4133} &74.54\%          &22.3\%                                     \\ 
    \bottomrule
  \end{tabular}
  \label{tab:Stage2Readout}
\end{table}

\begin{table}
 \caption{ MTCNN vs MTTST model parameters size}
  \centering
  \begin{tabular}{ll}
    \toprule
    \midrule
    Model & Params Size \\
    \midrule
    MTCNN      &\textbf{137K}               \\ 
    \midrule
    MTTST      &384K \\
    \bottomrule
  \end{tabular}
  \label{tab:ModelParams}
\end{table}

\section{Data Availability}
The datasets used in this study are proprietary and contain confidential e-Commerce data owned by our company. Therefore, they cannot be made publicly available. The availability of the model's source code is subject to the company's legal guidelines and ongoing internal audits.

\section{Conclusion}
In this paper, we proposed an MTCNN Behavioral Embedding Model for Transaction Fraud Detection, offering several innovative solutions to the outlined challenges.The architecture we propose utilizes a single-layer CNN with multi-scale kernels to efficiently extract transactional behaviors. This approach not only enhances computational scalability but also inherently incorporates expert knowledge from the payment risk domain. Furthermore, we combined the positional encoding in our architecture, as an effective remedy of overlooking the temporal order information for the conventional CNN. In addition, we incorporated a multitask learning framework to simultaneously address several correlated tasks associated with transaction fraud detection. This approach facilitates shared knowledge across all tasks and improves the model's generalization capabilities. Finally, we devised a strategy to effectively handle continuous variables in MTS data, introducing a scaling embedding method combined with categorical variable embeddings, making our domain data available in integration with most of the MTS methods. Experimental results from real-world industry data validated the effectiveness of our proposed model. We conducted experiments in 2 stages, they showed that MTCNN has competitive performance even compared to state-of-the-art Transformer solutions like TST, also with enhanced stability and robust generalization capabilities. 

Although these results are promising, we acknowledge the potential for further enhancement of our model. Future work will focus on integrating additional CNN and multitasking-related techniques to broaden the capabilities of our model. By continuously improving our methodology, we aim to advance the state-of-the-art in transaction fraud detection and ensure a safer and more trustworthy environment for online transactions.

\bibliography{references}

\end{document}